\def\year{2022}\relax
\documentclass[letterpaper]{article} %
\usepackage{aaai22}  %
\nocopyright
\usepackage{times}  %
\usepackage{helvet}  %
\usepackage{courier}  %
\usepackage[hyphens]{url}  %
\usepackage{graphicx} %

\usepackage{epsfig}
\usepackage{booktabs}
\usepackage{colortbl}
\usepackage{amsmath}
\usepackage{amssymb}
\usepackage{multicol}
\usepackage{multirow}
\usepackage{float}
\usepackage{placeins}
\usepackage{multirow}
\usepackage{url}

\usepackage{makecell}
\usepackage{subcaption}
\usepackage{cuted}
\usepackage[font={small}]{caption}
\usepackage{pifont}
\usepackage[dvipsnames]{xcolor}
\usepackage{balance}
\usepackage{epigraph}

\usepackage[pagebackref=true,breaklinks=true,colorlinks,bookmarks=false]{hyperref}

\usepackage{natbib}  %
\usepackage{caption} %
\DeclareCaptionStyle{ruled}{labelfont=normalfont,labelsep=colon,strut=off} %
\usepackage{algorithm}
\usepackage{algorithmic}

\usepackage{newfloat}
\usepackage{listings}
\floatstyle{ruled}
\newfloat{listing}{tb}{lst}{}
\floatname{listing}{Listing}
\title{MoCaNet: Motion Retargeting in-the-wild via Canonicalization Networks}

\author{
    Wentao Zhu,\equalcontrib\textsuperscript{\rm 1,2}
    Zhuoqian Yang,\equalcontrib\textsuperscript{\rm 3} 
    Ziang Di,\textsuperscript{\rm 4}
    Wayne Wu,\textsuperscript{\rm 2,3}\thanks{Corresponding author.}
    Yizhou Wang,\textsuperscript{\rm 1}
    Chen Change Loy\textsuperscript{\rm 5}
}
\affiliations{
    \textsuperscript{\rm 1} School of Computer Science, Peking University
    \textsuperscript{\rm 2} Shanghai AI Laboratory
    \textsuperscript{\rm 3} SenseTime Research \\
    \textsuperscript{\rm 4} Southeast University
    \textsuperscript{\rm 5} S-Lab, Nanyang Technological University \\
    wtzhu@pku.edu.cn, 
    yangzhuoqian@sensetime.com,
    213180794@seu.edu.cn,\\
    wuwenyan0503@gmail.com,
    yizhou.wang@pku.edu.cn,
    ccloy@ntu.edu.sg

}

\newcommand{\ie}{\textit{i}.\textit{e}., }
\newcommand{\eg}{\textit{e}.\textit{g}., }

\begin{document}

\maketitle

\begin{abstract}
We present a novel framework that brings the 3D motion retargeting task from controlled environments to in-the-wild scenarios. In particular, our method is capable of retargeting body motion from a character in a 2D monocular video to a 3D character without using any motion capture system or 3D reconstruction procedure. It is designed to leverage massive online videos for unsupervised training, requiring neither 3D annotations nor motion-body pairing information. The proposed method is built upon two novel canonicalization operations, structure canonicalization and view canonicalization. Trained with the canonicalization operations and the derived regularizations, our method learns to factorize a skeleton sequence into three independent semantic subspaces, \ie motion, structure, and view angle. The disentangled representation enables motion retargeting from 2D to 3D with high precision. Our method achieves superior performance on motion transfer benchmarks with large body variations and challenging actions. Notably, the canonicalized skeleton sequence could serve as a disentangled and interpretable representation of human motion that benefits action analysis and motion retrieval\footnote{Project page: \color{blue}{\url{https://yzhq97.github.io/mocanet}}}.
\end{abstract}

\section{Introduction}

3D motion retargeting aims at transferring one character's motion to a virtual 3D avatar. It is an important and challenging task in computer vision and computer graphics with a wide spectrum of applications in human-computer interaction~\cite{10.1145/3306346.3323034, 9196948} and augmented reality~\cite{8797856, 7523447}. Traditional approaches rely on motion capture (MoCap) systems~\cite{DBLP:conf/humanoids/OttLN08, DBLP:conf/icra/KoenemannBB14}, which are capable of capturing precise 3D motion but are limited to  sophisticated environments only available to film and game studios. This kind of setup is inaccessible for everyday users with in-the-wild usage scenarios, \eg on-device augmented reality. %

The challenge of in-the-wild motion retargeting lies in that videos recorded on mobile devices or downloaded from the Internet only provide 2D visual information (usually noisy), from which 3D motion needs to be inferred. The common practice is to estimate 3D pose first, then perform 3D-to-3D retargeting through Inverse Kinematics (IK). Nevertheless, the results of current 3D pose estimation~\cite{ci2019optimizing, pavllo20193d, Moon_2019_ICCV_3DMPPE, yang20183d} and model-based mesh reconstruction~\cite{lwb2019, kocabas2019vibe, Choi_2020_ECCV_Pose2Mesh} methods deteriorate severely under occlusions, large body variations and uncommon actions from unconstrained real-world videos, as discussed in~\cite{DBLP:conf/eccv/DongSZLZB20, kwanyee2019weakly3dpose, DBLP:journals/corr/abs-2104-13586}. The errors originated from 3D pose estimation can easily propagate to the IK stage and degrade the quality of retargeted motion.

\begin{figure}[t]
  \centering
  \includegraphics[width=0.9\linewidth]{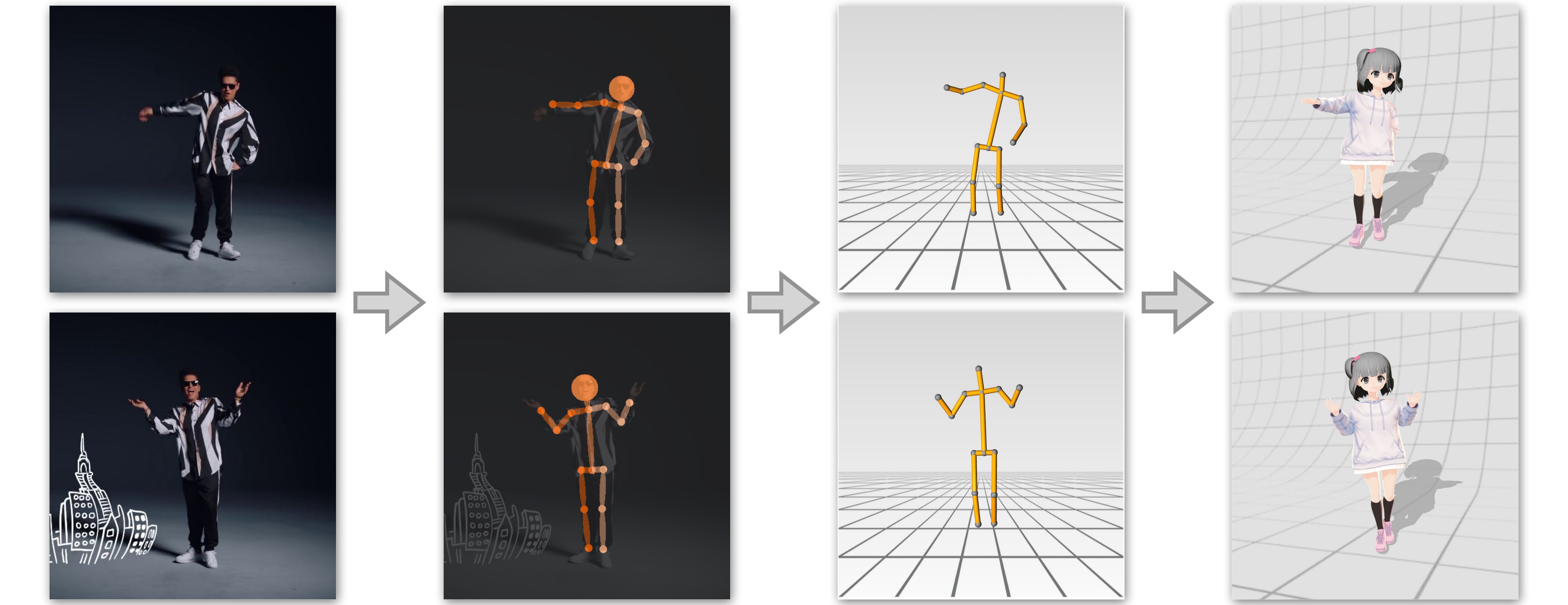}
  \caption{\textbf{3D Motion Retargeting in the Wild.} Motion is extracted from the video clip and retargeted to the virtual avatar. Our method extracts 2D skeleton sequences rather than 3D sequences from in-the-wild videos.}
  \label{fig:banner}
  \vspace{-0.25cm}
\end{figure}

In this work, we argue that the conventional way of performing IK after 3D estimation is not sufficiently robust and reliable for in-the-wild motion retargeting. To this end, we propose an end-to-end learnable model, \textit{Canonicalization Networks}, which allows us to bypass the error-prone direct 3D pose estimation and use an off-the-shelf 2D pose estimator to extract geometrical information that is more robust and reliable. Our method then takes the 2D skeleton sequences and generates 3D retargeted skeleton sequences, as shown in Fig.~\ref{fig:banner}. However, retargeting motion from 2D skeleton sequences is not a trivial task. Arbitrarily captured real-world videos contain large structural variations and diverse camera views. The same action, \eg standing up, may appear dramatically different in 2D under different views given different body structures. It is therefore meaningful and challenging to extract the unmixed motion in a high-level semantic modality.

\begin{figure}[t]
 \centering
 \includegraphics[width=\linewidth]{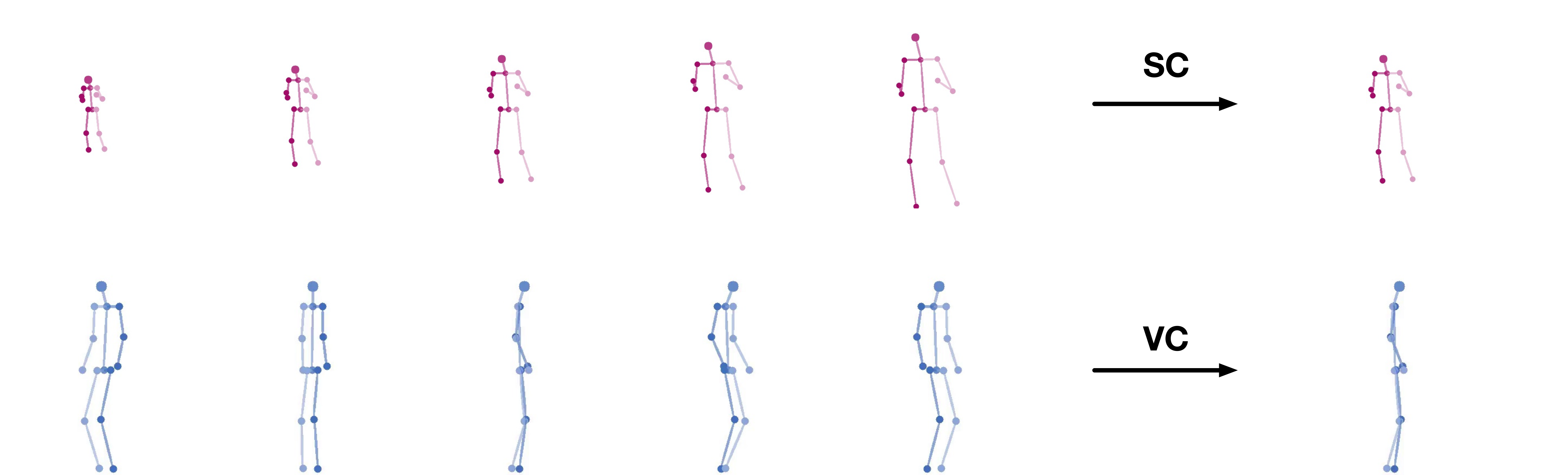}
 \caption{\textbf{Canonicalization Operations.} The first row shows the idea of \textit{Structure Canonicalization}, and the second row shows the idea of \textit{View Canonicalization}. Each skeleton represents an individual video clip.}
 \label{fig:canoidea}
\end{figure}

\begin{figure}[t]
 \centering
 \includegraphics[width=\linewidth]{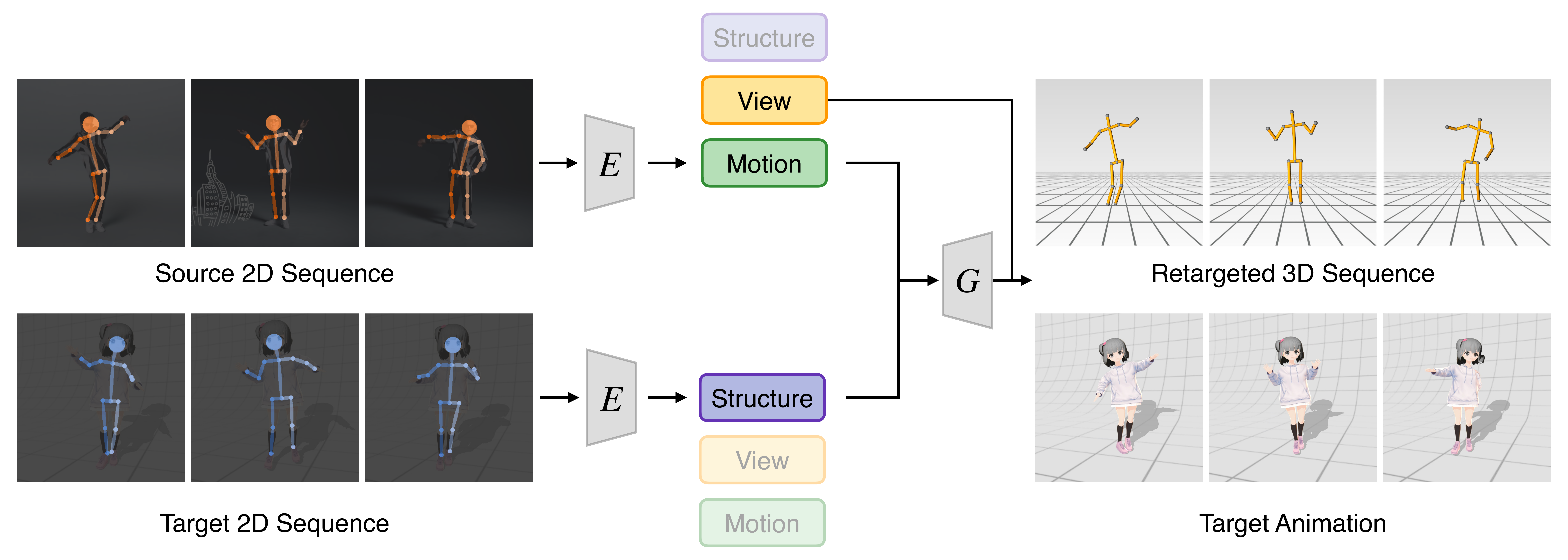}
 \caption{\textbf{Motion Retargeting Inference Pipeline.} The encoder, $E$, decomposes the input skeleton sequence into three latent codes. The decoder, $G$, takes the motion code from the source character and the structure code from the target character, then cast the decoded 3D skeleton sequence to the source view angle. This yields the retargeted 3D sequence, which could be used to animate the target character.}
 \label{fig:arch}
\end{figure}

We tackle the challenge through the novel notion of \textit{canonicalization and retargeting}. Canonicalization, by definition, aims at eliminating variations in a specific domain. In our context, we explore the canonicalization of structure and view angle from 2D skeleton sequences while keeping the motion unchanged.
Specifically, we design two parallel canonicalization operations, structure canonicalization and view canonicalization, as shown in Fig. \ref{fig:canoidea}. Structure canonicalization yields skeleton sequences with a uniform body structure, while the motion and other features are preserved. Similarly, view canonicalization provides skeleton sequences with the same view angle, casting different sequences to a uniform view. 
The idea behind is that a 2D skeleton sequence can be formulated as a product of three independent latent variables, namely, motion, structure, and view. By learning the canonicalization operations, the model should be able to disentangle the three meaningful factors from 2D skeleton sequences. It then allows us to freely recombine motion with arbitrary structures and view angles to generate the desired 3D output, as shown in Fig.~\ref{fig:arch}.

Based on these canonicalization operations, we formulate a set of novel self-supervised canonicalization losses, which randomly perturb a targeted factor (structure or view angle subspaces) and apply canonicalization in that space for both the original sequence and the manipulated sequence. Hence, the model can be trained on an extensive collection of human pose sequences extracted from Internet videos without any annotation, which notably increases the robustness and generalization of the model. %
Thanks to canonicalization and unsupervised training, the network is exposed to a wide range of variations. Such exposure and the requirement of restoring them to a specific canonical form (structure and view) allows the network to learn focusing on the inherent motion and omitting noisy information from the input sequence. This contributes to learning well-disentangled representations for precise motion retargeting.

The contributions of this work are three-fold: 1) We propose an end-to-end learnable model, \textit{Canonicalization Networks} to tackle the challenging in-the-wild 2D-to-3D motion retargeting problem. 2) We design an unsupervised learning approach based on the novel canonicalization operations, requiring neither paired data nor 3D annotations. Our approach circumvents the errors in monocular 3D estimation and benefits from large-scale web data for unsupervised training, which leads to improved performance demonstrated in our experiments. 3) Canonicalized skeleton sequences could serve as an interpretable representation of human motion, paving the way for related research and applications.

\section{Related Work}

\noindent
\textbf{Motion Retargeting}. 
Motion retargeting has been extensively studied in the area of computer vision and graphics. Classical motion retargeting methods~\cite{DBLP:conf/siggraph/HodginsP97, 10.1145/280814.280820, DBLP:journals/tog/TakK05, DBLP:conf/siggraph/LeeS99} mainly rely on simplified assumptions and hand-crafted kinematic constraints. In recent years, deep-learning-based approaches show promising results on obtaining human pose and action, which not only increases the availability of motion data, but also inspires motion retargeting research with deep neural networks. \citeauthor{NKN} proposed to capture high-level motion with Forward Kinematics (FK) layers in recurrent neural networks. \citeauthor{lim2019pmnet} further optimized the retargeting precision by disentangling pose and movement. Skeleton-Aware Networks~\cite{SANDMR2020} could automatically adapt to different skeleton topologies. Nevertheless, the above methods all require high-precision 3D motion (\eg quaternions), which takes expensive motion capture systems or complex optimization in practice. Otherwise, estimating 3D motion from monocular RGB video could be error-prone~\cite{DBLP:conf/eccv/DongSZLZB20, kwanyee2019weakly3dpose, DBLP:journals/corr/abs-2104-13586}.

Several works have also explored to retarget motion from 2D inputs that are more accessible to everyday use cases. \citeauthor{DBLP:journals/cgf/AbermanSLLCC19} proposed a two-branch framework with part confidence map as 2D pose representations. \citeauthor{EDN} designed a global pose normalization to handle different body structures. \citeauthor{LCM} proposed a supervised learning approach to retarget motion in 2D. It requires the exact same motion performed by different characters at different view angles as training data. TransMoMo~\cite{yang2020transmomo} explored unsupervised motion retargeting via invariance properties. These methods all generate 2D label maps for further rendering via image-to-image translation methods. Therefore, they can not be readily applied to driving 3D characters. They only consider accuracy for 2D joint positions and assume static view angle. Meanwhile, our canonicalization-based approach enables applying regularization on 3D poses directly and training with time-varying explicit 3D view.

In our framework, 2D poses are used as input and our method produces 3D skeletons as output. Therefore, our approach can exploit user-friendly monocular RGB videos with the help of robust 2D pose estimation algorithms~\cite{cao2018openpose, bulat2020toward, DBLP:conf/cvpr/ArtachoS20}. In addition, the 3D skeleton output allows combination with state-of-the-art neural rendering~\cite{wang2018vid2vid, wang2019fewshotvid2vid} and avatar creation~\cite{saito2020pifuhd, DBLP:conf/eccv/DengLJPH0T20, alldieck2018video, weng2019photo} techniques. Our flexible approach thus provides a viable alternative to the traditional 3D reconstruction+IK pipeline.

\vspace{0.1cm}
\noindent
\textbf{Representation Disentanglement}.
Disentangling latent variables from observations has been a fundamental topic in machine learning. Previous approaches~\cite{KingmaMRW14, NIPS2016_ef0917ea} use labeled data to separate class-dependent and class-independent features. A vast literature focuses on unsupervised learning of disentangled representations with generative models \cite{chen2016infogan, DBLP:conf/nips/DentonB17, DBLP:conf/iclr/VillegasYHLL17, DBLP:conf/cvpr/Tulyakov0YK18, higgins2017beta, DRIT, huang2018munit}. A recent line of work explored view-angle-agnostic motion representations aimed at improving motion recognition performance \cite{liu2021normalized, nie2020unsupervised, zhao2021learning}. In this work, we achieve disentanglement of motion, body-structure and view-angle via unsupervised canonicalization.

\vspace{0.1cm}
\noindent
\textbf{3D Canonicalization}.
Several different methods have been proposed for automatically computing the canonical form from a 3D pose or mesh. \cite{3dor.20151063, PICKUP20152500, PICKUP201817, DBLP:journals/cvm/PickupSRM16} propose to learn pose-neutral shape of any non-rigid meshes for 3D shape retrieval. \cite{3dor.20151057} further improves the method with multidimensional scaling. \citeauthor{DBLP:journals/ijcv/LianGX13} present a feature-preserved canonical form that directly deforms the 3D models. Body reconstruction methods~\cite{saito2020pifuhd, PTF:CVPR:2021} sometimes involve the canonical pose space for handling pose-dependent deformations. In those papers, canonicalization refers to \textit{pose canonicalization}, that is, ``transforming to a standard pose''. In this work, we explore the canonicalization of different factors, namely structure and view angle, while keeping the motion unchanged. C3DPO~\cite{C3DPO} designed a rotation-invariant canonical shape for structure reconstruction. CanonPose~\cite{Wandt2021Canonpose} learns a canonical camera view and respective camera rotations, but it relies on multiple synchronized cameras for training. We design canonicalization operations for both structure and view angle, which are applied to the entire action sequence rather than a single pose. Based on the designed operations, our model can be trained without data pairing.

\section{Methodology}

\subsection{Overview}

Our solution to in-the-wild 3D motion retargeting uses 2D skeleton sequences for training and inference. Specifically, we extract pose sequences from unannotated web videos with an off-the-shelf human pose estimator \cite{alp2018densepose, Detectron2018}. 
We devise a 2D-to-3D autoencoder that factorizes the input 2D human pose sequence into three independent spaces, namely character-agnostic 3D motion, body structure of the character and view angle. Notably, we formulate the view angle to be explicit and dynamic, which allows for direct manipulation. 3D motion retargeting is then achieved through decoding a recombination of motion and structure encoded from different sources.

We design a training scheme that simultaneously achieves (i) 3D geometry inference and (ii) disentangled motion representation learning. The two goals are tightly coupled and mutually beneficial. The first goal is addressed via a 2D-to-3D AutoEncoder trained with adversarial learning~\cite{chen2019unsupervised}. The second challenge is addressed via the proposed self-supervised canonicalization.

\begin{figure}[t]
  \centering
  \includegraphics[width=\linewidth]{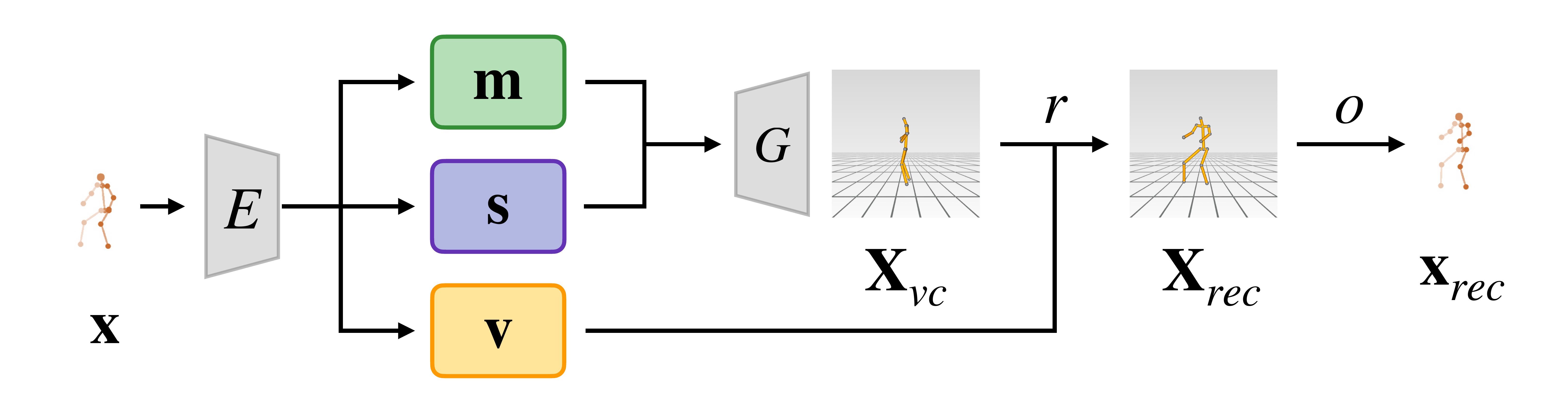}
  \caption{\textbf{2D-to-3D AutoEncoder Structure.} The encoder $E$ encodes an input 2D skeleton sequence $\mathbf{x}$ as character-agnostic motion $\mathbf{m}$, structure $\mathbf{s}$ and view $\mathbf{v}$. The decoder $G$ takes as input the concatenation of $\mathbf{m}$ and $\mathbf{s}$ and outputs a 3D skeleton sequence $\mathbf{X}_{vc}$ in canonical view. The reconstructed 3D sequence in original camera view is obtained by rotating $\mathbf{X}_{vc}$ with the encoded view angles $\mathbf{v}$, \ie $\mathbf{X}_{rec} = r(\mathbf{X}_{vc}, g(\mathbf{v}))$. The function $g(\cdot)$ converts the view $\mathbf{v}$ to rotation matrices. The reconstructed 2D sequence $\mathbf{x}_{rec}$ is obtained after an orthographic projection, \ie $\mathbf{x}_{rec}=o(\mathbf{X}_{rec})$. }
  \label{fig:ae}
\end{figure}

\subsection{2D-to-3D AutoEncoder}
\label{sec_ae}

 \begin{figure*}[t]
   \centering
   \includegraphics[width=\linewidth]{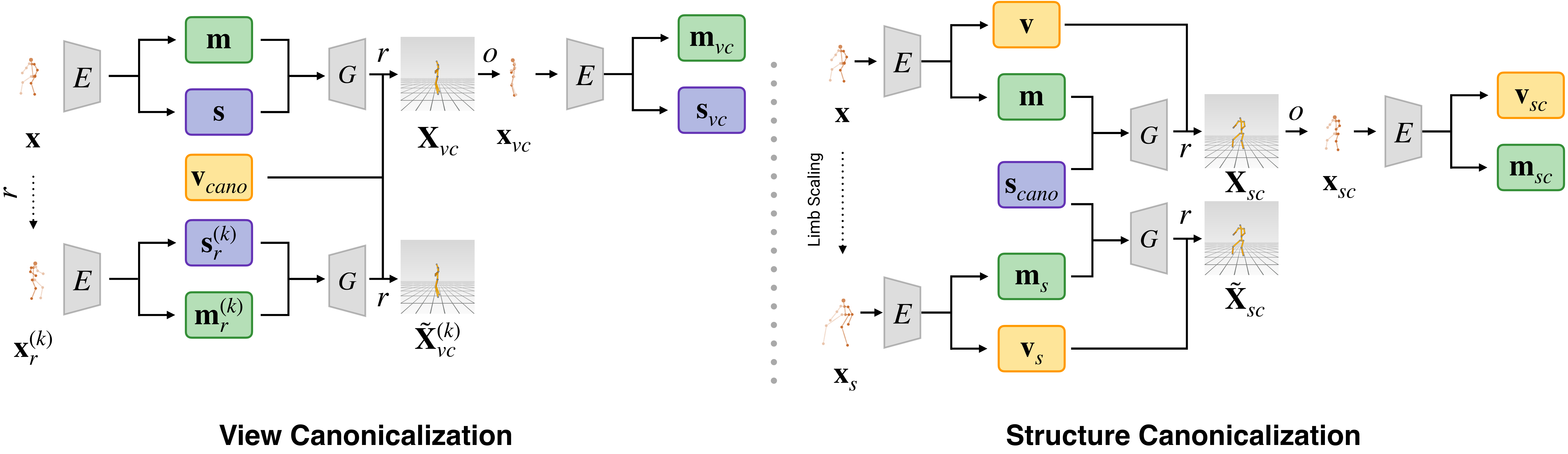}
   \caption{\textbf{Canonicalization.} (a) View canonicalization $\phi_{vc}(\cdot)$ is to restore canonical view angle $\mathbf{v}_{cano}$ for a skeleton sequence. During training, view canonicalization is applied to both the input 2D sequence $\mathbf{x}$ and its randomly rotated versions $\mathbf{x}_r^{(k)}$, \ie $\mathbf{X}_{vc}=\phi_{vc}(\mathbf{x}), \tilde{\mathbf{X}}_{vc}^{(k)}=\phi_{vc}(\mathbf{x}_r^{(k)})$.  (b) Structure canonicalization $\phi_{sc}(\cdot)$ is to store standard body structure $\mathbf{s}_{cano}$ for a skeleton sequence. During training, structure canonicalization is applied to both the input 2D sequence $\mathbf{x}$ and its randomly limb-scaled version $\mathbf{x}_s$, \ie $\mathbf{X}_{sc}=\phi_{sc}(\mathbf{x}), \tilde{\mathbf{X}}_{sc}=\phi_{sc}(\mathbf{x}_s)$.}
   \label{fig:cano}
 \end{figure*}

As shown in Fig. \ref{fig:ae}, the proposed 2D-to-3D AutoEncoder is a modular encoder-decoder network, where the encoding part decomposes an input 2D skeleton sequence $\mathbf{x} \in \mathbb{R} ^ {2N \times T}$ into three spaces: motion, structure and view angle, while the decoding part takes any combination of motion and structure and decodes a 3D skeleton sequence. Here, $N$ denotes the number of body joints and $T$ denotes the sequence length. The decoded 3D skeleton sequence should reproduce the input when projected back to 2D while remain plausible when it is viewed at different angles. This serves as the foundation of the representation learning framework and allows us to manipulate the data for canonicalization training.

\paragraph{Encoders.} The encoded representation of the three spaces are defined as follows: Motion is represented as a temporal sequence of latent vectors $E_m(\mathbf{x})=\mathbf{m} \in \mathbb{R}^{M \times C_m}$ where $M$ is the encoded length and $C_m$ is the number of latent channels. 
The motion encoder uses several 1-D temporal convolutional layers to extract this information, yielding the encoded length $M = T / 8$. The structure encoder has a similar network structure $E_s(\mathbf{x})=\mathbf{s}\in \mathbb{R}^{M \times C_s}=[\mathbf{s}^{(1)}, \mathbf{s}^{(1)}, ..., \mathbf{s}^{(M)}]$. This can be interpreted as performing multiple structure estimations in sliding windows. Since the structure does not change with time, we use a temporal max pooling to aggregate information from these estimations to get the final structure: $\bar{E_s}(\mathbf{x})=\bar{\mathbf{s}}=\text{maxpool}(\mathbf{s}), \bar{\mathbf{s}} \in \mathbb{R}^{C_s}$.
We assume view angle to be dynamic, therefore it is encoded as a temporal sequence of explicit 3D rotations. Instead of the commonly used Euler angles or quaternions, we choose to use a continuous 6D representation, which is better suited for deep learning regression \cite{zhou2019continuity}. Specifically, $E_v(\mathbf{x})=\mathbf{v} \in \mathbb{R}^{T \times 6}$, \ie $\mathbf{v}$ is a sequence of 6D rotation vectors, each rotation vector $\mathbf{v}_i$ is responsible for rotating one frame. $\mathbf{v}$ can be uniquely translated to a sequence of rotation matrices $\mathbf{\Omega} \in \mathbb{R}^{T \times 3 \times 3} = [\boldsymbol{\omega}^{(1)}, \boldsymbol{\omega}^{(2)}, ..., \boldsymbol{\omega}^{(T)}], \boldsymbol{\omega}^{(i)} \in SO3$ using a function: $g(\mathbf{v}) = \boldsymbol{\Omega}$.

\paragraph{Decoder.} As shown in Fig. \ref{fig:ae}, the decoder takes the concatenation of the motion and structure as input to decode a 3D skeleton sequence $G(\mathbf{m}, \mathbf{s}) = \mathbf{X}_{vc} \in \mathbb{R}^{3N \times T}$. The decoded sequence is assumed to be in the canonical view angle. To reproduce the input 2D sequence, the output 3D skeleton sequence $\mathbf{X}_{vc}$ is rotated using the encoded view angles $\mathbf{v}$ and then projected to 2D, \ie $\mathbf{X}_{rec}=r(\mathbf{X}_{vc}, g(\mathbf{v}))$, $\mathbf{x}_{rec} = o(\mathbf{X}_{rec})$. The function $r(\mathbf{X}, \mathbf{\Omega})$ is one which rotates a 3D skeleton sequence $\mathbf{X}$ with a sequence of SO3 rotation matrices $\boldsymbol{\Omega}$. $o(\mathbf{X})$ stands for orthographic projection.

\paragraph{3D Self-Supervision.} \label{sec:3dss} Since 3D ground-truth is not available, the ill-posed 2D-to-3D lifting is achieved via supervision in 2D space. Specifically, we use (i) an $L1$ reconstruction loss to ensure that the reprojected 2D skeleton sequence reproduces the input and (ii) an adversarial loss to ensure the reconstructed $3D$ skeleton, when viewed at random angles, still lies within the distribution of real $2D$ skeleton sequences. Specifically, the $L1$ reconstruction loss is defined as
\begin{equation}
    \mathcal{L}_\text{rec} = \left| \mathbf{x} - \mathbf{x}_{rec} \right|,
\end{equation}
where $\mathbf{x}$ is the input 2D sequence and $\mathbf{x}_{rec}$ is the sequence reconstructed by the autoencoder. To calculate the adversarial loss, we rotate the lifted 3D sequence with some random view angles. We generate $K$ versions of rotated sequences $\mathbf{X}_r^{(k)} = r(\mathbf{X}_{rec}$, $\boldsymbol{\Omega}_k),  \mathbf{x}_r^{(k)} = o(\mathbf{X}_r^{(k)})$. Note that the rotation matrices are time-varying, and in practice we keep them temporally smooth by interpolation. A discriminator $D$ is used for the adversarial training. The discriminator tries to distinguish between real and reprojected 2D sequences while the AutoEncoder tries to make its 3D output indistinguishable regardless of viewing angle. This process eliminates the ambiguity in the 2D-to-3D mapping by penalizing unusual structures in the reprojected views \cite{chen2019unsupervised}. The adversarial loss is defined as
\begin{equation}
    \mathcal{L}_\text{adv} = \sum_{k=1}^K \mathbb{E}_{\mathbf{x} \sim p_\mathbf{x}} [ \log D(\mathbf{x}) +  \log(1-D(\mathbf{x}_{r}^{(k)})]
    \label{eq:loss_adv}
\end{equation}

\subsection{Canonicalization}
\label{sec_cano}
Canonicalization aims at restoring canonicality in one of the spaces while leaving the other two unchanged, a process that enforces the independence of one space against the other two. We propose view canonicalization (restore canonical view for any 3D skeleton sequence) and structure canonicalization (restore canonical body structure for any 3D skeleton sequence). We randomly manipulate the 3D sequence in only one of the three spaces and apply canonicalization in that space for both the original sequence and the manipulated sequence. This process gives us several variants that can be used to derive the supervision signals.

\paragraph{View Canonicalization.}
\label{sec:vc}

View canonicalization $\phi_{vc}(\mathbf{x})$ restores canonical view for any skeleton sequence, \ie performing view canonicalization on figures with arbitrary orientation should yield figures facing the same direction, while leaving the motion and body structure unchanged. This is achieved by reconstructing the 3D sequence with the decoder and then rotating it to canonical view, \ie $\phi_{vc}(\mathbf{x})=r(G(E_m(\mathbf{x}), E_s(\mathbf{x})), g(\mathbf{v}_{cano}))$. The canonical view $\mathbf{v}_{cano}$ is defined as corresponding to the identity matrix, \ie no rotation. Thus, the output of the decoder is assumed to be in a canonical view, see Fig.~\ref{fig:cano} (a) for an example.

Recall in the previous section that we generated 2D reprojections $\mathbf{x}_r^{(k)}$ of the input sequence $\mathbf{x}$. For training, we perform view canonicalization on both the input and the reprojected versions. Since $\mathbf{x}$ and $\mathbf{x}_r^{(k)}$ contain the same motion and structure and the only variable is the view angle, the results of view canonicalization should be identical. From this property we derive the view canonicalization loss in skeleton space:
\begin{equation}
\mathcal{L}^\text{vc}_X
= \sum_{k=1}^{K} \left| \mathbf{X}_{vc} - \tilde{\mathbf{X}}_{vc}^{(k)} \right|,
\end{equation}
where $\mathbf{X}_{vc}=\phi_{vc}(\mathbf{x})$ and $\tilde{\mathbf{X}}_{vc}^{(k)}=\phi_{vc}(\mathbf{x}_r^{(k)})$. Also, since rotation and view canonicalization does not modify information in the motion and structure space, the re-encoded motion and structure should stay the same. These properties give us the view canonicalization loss in feature space.
\begin{equation}
\mathcal{L}^\text{vc}_m = \left| \mathbf{m} - \mathbf{m}_{vc} \right| + \sum_{k=1}^K \left| \mathbf{m} - \mathbf{m}_r^{(k)} \right|,
\end{equation}
\begin{equation}
\mathcal{L}^\text{vc}_s = \left| \mathbf{s} - \mathbf{s}_{vc} \right| + \sum_{k=1}^K \left| \mathbf{s} - \mathbf{s}_r^{(k)} \right|,
\end{equation}
where $\mathbf{m}$ and $\mathbf{s}$ are the motion and structure encoded from the input sequence $\mathbf{x}$. Together, the \textit{view canonicalization loss} is defined as
\begin{equation}
\mathcal{L}_{vc} = \lambda^{vc}_X \mathcal{L}^\text{vc}_X + \lambda^{vc}_m \mathcal{L}^\text{vc}_m + \lambda^{vc}_s \mathcal{L}^\text{vc}_s .
\end{equation}

\paragraph{Structure Canonicalization.}
\label{sec:sc}

Structure canonicalization restores a standard body structure for any skeleton sequence while leaving the motion and view angle unchanged. During training, we estimate a centroid of the structure space and define it as the canonical structure $\mathbf{s}_{cano}$. Structure canonicalization is defined as $\phi_{sc}(\mathbf{x}) = r\left( G(E_m(\mathbf{x}), \mathbf{s}_{cano}), E_v(\mathbf{x}) \right)$. The pipeline for structure canonicalization training is shown in Fig.~\ref{fig:cano} (b).

To achieve self-supervision, a technique called limb-scaling \cite{yang2020transmomo}, \ie randomly shortening or extending the length of the limbs in the input 2D sequence $\mathbf{x}$, is used to obtain a sequence $\mathbf{x}_s$, which has the same motion and view angle but a different body structure. We then perform structure canonicalization on both sequences, \ie $\mathbf{X}_{sc}=\phi_{sc}(\mathbf{x})$, $\tilde{\mathbf{X}}_{sc}=\phi_{sc}(\mathbf{x}_s)$. By definition, this pair of canonicalization results should be the same, from which we derive the structure canonicalization loss in the skeleton space:
\begin{equation}
\mathcal{L}^\text{sc}_X
= \left| \mathbf{X}_{sc} - \tilde{\mathbf{X}}_{sc} \right|.
\end{equation}

Since limb-scaling and structure canonicalization does not modify information in the other two spaces, the re-encoded motion and view angle should stay the same. These properties give us the structure canonicalization loss in feature space.
\begin{equation}
\mathcal{L}^\text{sc}_m = \left| \mathbf{m} - \mathbf{m}_{sc} \right| + \left| \mathbf{m} - \mathbf{m}_s \right|,
\end{equation}
\begin{equation}
\mathcal{L}^\text{sc}_v = \left| \mathbf{v} - \mathbf{v}_{sc} \right| + \left| \mathbf{v} - \mathbf{v}_s \right|.
\end{equation}
Together, the \textit{structure canonicalization loss} is defined as
\begin{equation}
\mathcal{L}_{sc} = \lambda^{sc}_X \mathcal{L}^\text{sc}_X + \lambda^{sc}_m \mathcal{L}^\text{sc}_m + \lambda^{sc}_v \mathcal{L}^\text{sc}_v.
\end{equation}

\subsection{Total Loss}

The total training loss is given as
\begin{align}
\mathcal{L} = \lambda_{rec} \mathcal{L}_{rec} + \lambda_{adv} \mathcal{L}_{adv} + \mathcal{L}_{vc} + \mathcal{L}_{sc}. 
\end{align}

We back-propagate $\mathcal{L}$ and train the whole network in an end-to-end fashion.

\section{Experiments}

In this section, we first present the implementation details. We then evaluate and compare the motion retargeting performance on both in-the-wild and synthetic data. Next, we investigate the effects of canonicalization on the learned representations. We also conduct an ablation study to examine the effectiveness of each module. We include the details of the neural network and the metrics in the appendix.

\subsection{Implementation Details}
\paragraph{Dataset.}
For in-the-wild training, we use the Solo-Dancer\cite{yang2020transmomo} dataset which is a collection of YouTube videos featuring a single dancer in the scene. The dataset contains $8$ categories of $337$ videos. The total length of the videos add up to $11.5$ hours. We then used an off-the-shelf 2D keypoints detector \cite{alp2018densepose} to extract keypoints frame-by-frame to be used as our training data. We also perform the proposed unsupervised training pipeline on the synthetic Mixamo dataset \cite{mixamo} in order to quantitatively measure the transfer results with ground truth and baseline methods. The training set comprises of $32$ characters, each character has $800$ sequences and a total of $1.2$ hours for each character. During evaluation, we test the models on a held-out partition with $4$ new characters and $64$ new motions.

\paragraph{Training.}
Our canonicalization networks are trained for $200000$ steps with batch size $64$ and learning rate $0.0001$ using an Adam~\cite{kingma2014adam} optimizer. The full model is end-to-end trained with all the proposed loss terms. The weights of the loss terms are given as follows: $\lambda_\text{rec}=15, \lambda_\text{adv}=2, \lambda_\text{trip}=10, \lambda^{vc}_X=\lambda^{sc}_X=5, \lambda^{vc}_m=\lambda^{sc}_m=\lambda^{vc}_s=\lambda^{sc}_v=2$. For view canonicalization, we generate $K=3$ randomly rotated versions of the reconstructed 3D sequence. For limb-scaling in structure canonicalization, we use global and local scaling factors randomly sampled from a uniform distribution in the range $[0.5, 2]$. These parameters are empirically determined.  All the experiments are implemented on a Linux Ubuntu machine with \text{NVIDIA 1080Ti} GPU using PyTorch 1.4~\cite{NEURIPS2019_9015}. The model takes $3$GB of GPU memory and $8$ hours to train on a single GPU. We set all the random seeds to be $123$.

\begin{table*}[t]
\caption{\textbf{2D-to-3D (Synthetic) Motion Retargeting Comparison.} We calculate the 3D Mean Square Error (MSE $\times10^{-2}$) and 3D Mean Per Joint Position Error (MPJPE $\times10^{-1}$). Both MSE and MPJPE are root-relative and normalized by the target character's body height, following the practice of \cite{LCM}. The test characters are shown in Fig. \ref{fig:mixamo}.} 

\begin{center}
\resizebox{1.0\linewidth}{!}{
\small
\begin{tabular}{l|cccccccccccccc}
\Xhline{1.2pt}
Methods                      & \multicolumn{2}{c|}{3DGT+IK}                         & \multicolumn{2}{c|}{Ours}                                      & \multicolumn{2}{c|}{Ours (wild)}                                 & \multicolumn{2}{c|}{TransMoMo}                                 & \multicolumn{2}{c|}{LCN+IK}                                       & \multicolumn{2}{c|}{Pose2Mesh+IK}                                 & \multicolumn{2}{c}{VideoPose3D+IK}                               \\
Metrics                      & \multicolumn{1}{c}{$\text{MSE}$} & \multicolumn{1}{c|}{$\text{MPJPE}$} & \multicolumn{1}{c}{$\text{MSE}$}       & \multicolumn{1}{c|}{$\text{MPJPE}$}     & \multicolumn{1}{c}{$\text{MSE}$}       & \multicolumn{1}{c|}{$\text{MPJPE}$}     & \multicolumn{1}{c}{$\text{MSE}$}       & \multicolumn{1}{c|}{$\text{MPJPE}$}     & \multicolumn{1}{c}{$\text{MSE}$}       & \multicolumn{1}{c|}{$\text{MPJPE}$}     & \multicolumn{1}{c}{$\text{MSE}$}       & \multicolumn{1}{c|}{$\text{MPJPE}$}     & \multicolumn{1}{c}{$\text{MSE}$}       & \multicolumn{1}{c}{$\text{MPJPE}$}                         \\
\Xhline{1.2pt}
Andro. $\leftrightarrow$ P. Hulk     & 0.049 & 0.325 & \cellcolor[HTML]{FFFFFF}0.799 & \cellcolor[HTML]{FFFEFE}1.221 & \cellcolor[HTML]{FFFFFF}0.792 & \cellcolor[HTML]{FFFEFE}1.215 & \cellcolor[HTML]{FFFDF5}1.246 & \cellcolor[HTML]{FCEFED}1.563 & \cellcolor[HTML]{FFF8E4}1.986 & \cellcolor[HTML]{F9DEDC}1.925 & \cellcolor[HTML]{FFF3D1}2.822 & \cellcolor[HTML]{F5CAC7}2.361 & \cellcolor[HTML]{FFF1C9}3.198 & \cellcolor[HTML]{F4C5C1}2.469 \\
Andro. $\leftrightarrow$ S. Granny    & 0.042 & 0.294 & \cellcolor[HTML]{FFFFFC}0.948 & \cellcolor[HTML]{FEFAF9}1.320 & \cellcolor[HTML]{FFFFFC}0.945 & \cellcolor[HTML]{FEFAF9}1.320 & \cellcolor[HTML]{FFFAEC}1.668 & \cellcolor[HTML]{FAE2E0}1.830 & \cellcolor[HTML]{FFF8E5}1.973 & \cellcolor[HTML]{FAE0DE}1.881 & \cellcolor[HTML]{FFF5D9}2.504 & \cellcolor[HTML]{F7D4D1}2.159 & \cellcolor[HTML]{FFE499}5.309 & \cellcolor[HTML]{EEA69F}3.173 \\
Andro. $\leftrightarrow$ TY               & 0.016 & 0.168 & \cellcolor[HTML]{FFFFFE}0.853 & \cellcolor[HTML]{FFFFFF}1.206 & \cellcolor[HTML]{FFFFFF}0.832 & \cellcolor[HTML]{FFFFFF}1.191 & \cellcolor[HTML]{FFFBF0}1.497 & \cellcolor[HTML]{FBE7E6}1.725 & \cellcolor[HTML]{FFFAE9}1.769 & \cellcolor[HTML]{FBE7E5}1.730 & \cellcolor[HTML]{FFF9E6}1.910 & \cellcolor[HTML]{FAE0DE}1.878 & \cellcolor[HTML]{FFECB6}4.044 & \cellcolor[HTML]{F3BCB7}2.678 \\
P. Hulk $\leftrightarrow$ S. Granny & 0.131 & 0.507 & \cellcolor[HTML]{FFFFFC}0.955 & \cellcolor[HTML]{FEF9F9}1.334 & \cellcolor[HTML]{FFFFFC}0.951 & \cellcolor[HTML]{FEF9F9}1.333 & \cellcolor[HTML]{FFF8E3}2.057 & \cellcolor[HTML]{F8D8D5}2.072 & \cellcolor[HTML]{FFF7DF}2.235 & \cellcolor[HTML]{F8DAD8}2.013 & \cellcolor[HTML]{FFF1C9}3.182 & \cellcolor[HTML]{F4C4C0}2.491 & \cellcolor[HTML]{FFDD81}6.390 & \cellcolor[HTML]{E98B83}3.760 \\
P. Hulk $\leftrightarrow$ TY            & 0.064 & 0.371 & \cellcolor[HTML]{FFFFFE}0.874 & \cellcolor[HTML]{FFFEFE}1.230 & \cellcolor[HTML]{FFFFFE}0.853 & \cellcolor[HTML]{FFFEFE}1.220 & \cellcolor[HTML]{FFF9E8}1.837 & \cellcolor[HTML]{F9DCDA}1.965 & \cellcolor[HTML]{FFF7E1}2.135 & \cellcolor[HTML]{F9DDDB}1.942 & \cellcolor[HTML]{FFF5D8}2.544 & \cellcolor[HTML]{F6D0CD}2.235 & \cellcolor[HTML]{FFE395}5.471 & \cellcolor[HTML]{EC9A93}3.429 \\
S. Granny $\leftrightarrow$ TY           & 0.024 & 0.209 & \cellcolor[HTML]{FFFFFC}0.927 & \cellcolor[HTML]{FFFCFC}1.266 & \cellcolor[HTML]{FFFFFD}0.912 & \cellcolor[HTML]{FFFDFC}1.256 & \cellcolor[HTML]{FFF7E0}2.166 & \cellcolor[HTML]{F7D4D1}2.147 & \cellcolor[HTML]{FFF7E1}2.141 & \cellcolor[HTML]{F9E0DE}1.886 & \cellcolor[HTML]{FFF6DD}2.313 & \cellcolor[HTML]{F8D7D5}2.073 & \cellcolor[HTML]{FFD666}7.542 & \cellcolor[HTML]{E67C73}4.077 \\
OVERALL                      & 0.056 & 0.316 & \cellcolor[HTML]{FFFFFD}0.891 & \cellcolor[HTML]{FFFCFC}1.261 & \cellcolor[HTML]{FFFFFE}0.878 & \cellcolor[HTML]{FFFDFC}1.253 & \cellcolor[HTML]{FFFAEA}1.750 & \cellcolor[HTML]{F9E0DE}1.888 & \cellcolor[HTML]{FFF8E3}2.046 & \cellcolor[HTML]{F9DFDD}1.899 & \cellcolor[HTML]{FFF5D8}2.552 & \cellcolor[HTML]{F7D2CE}2.204 & \cellcolor[HTML]{FFE499}5.329 & \cellcolor[HTML]{EDA19B}3.272 \\
\Xhline{1.2pt}
\end{tabular}
}
\end{center}
\label{tab:mse}
\end{table*}

\begin{figure}[t]
  \centering
  \includegraphics[width=0.7\linewidth]{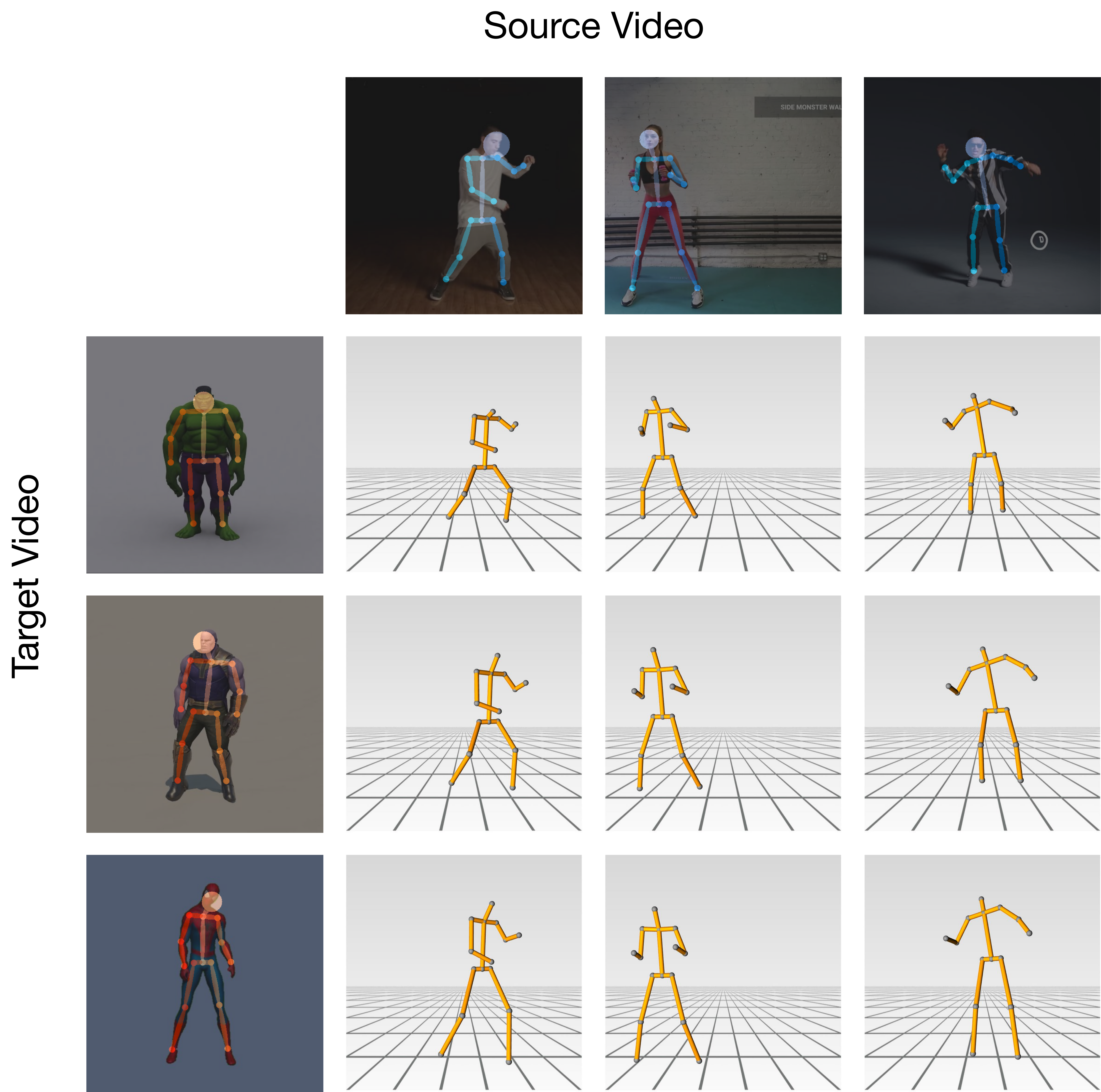}
  \caption{\textbf{3D Motion Retargeting in-the-Wild.} The uppermost row gives the source videos, and the leftmost column gives the target videos. For each source-target combination, the network transfers the motion from source to target and produces 3D results.}
  \label{fig:retargeting}
\end{figure}

\subsection{Motion Retargeting}

Figure \ref{fig:retargeting} shows the motion retargeting results of our method. Each output skeleton is conditioned on the motion and view code from the source character and the structure code from the target character. The results demonstrate that our method could transfer the motion naturally and realistically to the target body. In addition, we compare our method with the state-of-the-art methods. To implement the conventional 3D+IK pipeline, we use several SOTA 3D pose estimation and mesh reconstruction methods as detailed below.

\begin{figure}[h]
  \centering
  \includegraphics[width=\linewidth]{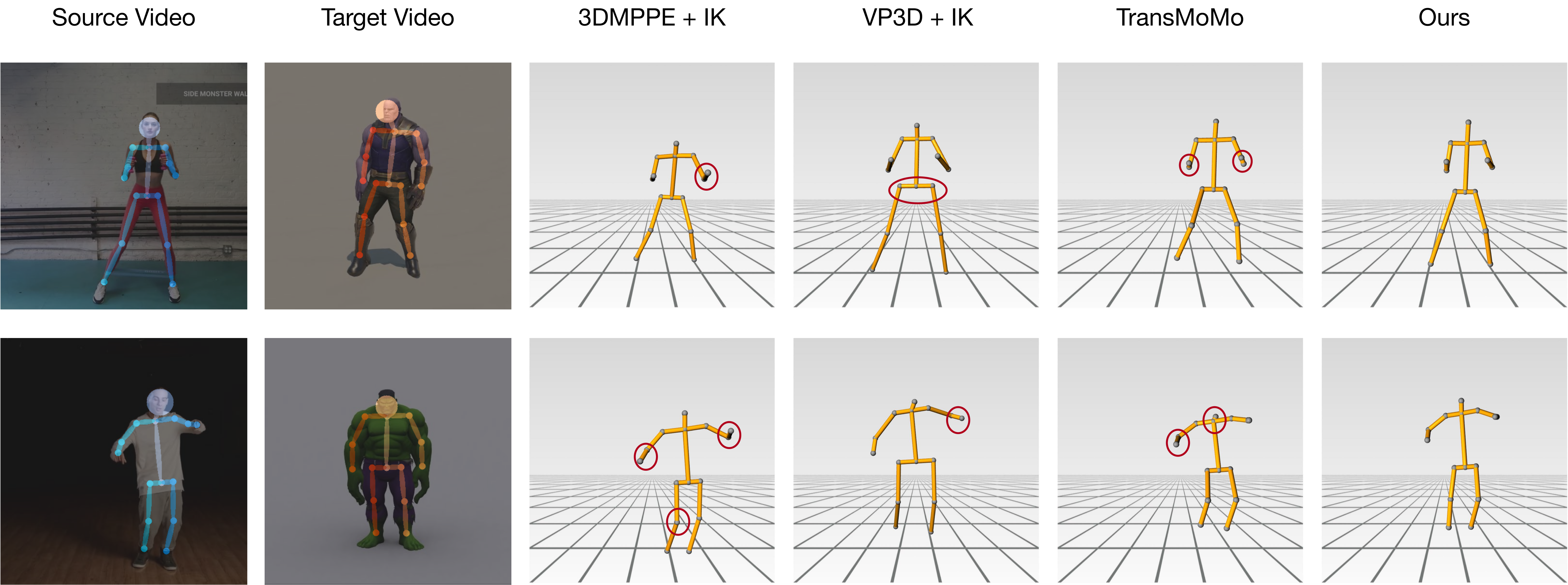}
  \caption{\textbf{Video-to-3D (In-the-Wild) Motion Retargeting Comparison.} The compared methods retarget the motion from source videos to target videos in the form of 3D skeleton sequences. Please zoom in for details, especially those circled with red that shows the erroneous results produced by existing methods.}
  \label{fig:comparison}
  \vspace{-0.35cm}
\end{figure}

\subsubsection{Video-to-3D (In-the-Wild) Comparison.} We first test under the in-the-wild setting where the source and target are given by dancing videos downloaded from the Internet, and the results are shown in Fig.~\ref{fig:comparison}. VideoPose3D~\cite{pavllo20193d} takes a 2D skeleton sequence and lifts it to 3D with temporal convolutions. 3DMPPE~\cite{Moon_2019_ICCV_3DMPPE} regresses 3D pose from the 2D image directly. In the traditional two-stage pipeline, errors in the 3D pose estimation stage are accumulated and enlarged in the IK stage. Therefore, these methods are highly reliant on the quality of 3D pose estimation, which tends to fail due to occlusions or short 2D projections. TransMoMo~\cite{yang2020transmomo} produces an intermediate 3D result before projection. However, it is designed to generate 2D label maps for further rendering, so they only consider 2D supervision with an oversimplified static view assumption. Although its 2D projection results are plausible, multiple artifacts can be found in 3D, \eg wrong direction, temporally twisted moves.

\begin{figure}[t]
  \centering
  \includegraphics[width=\linewidth]{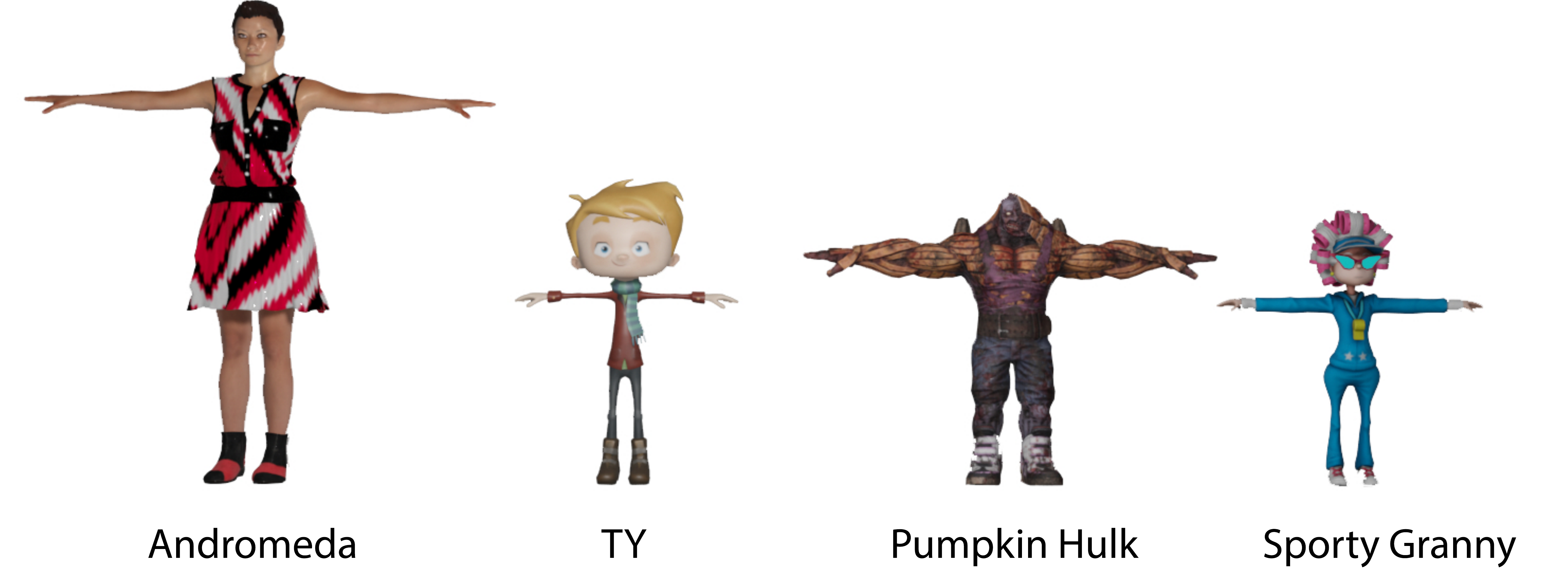}
  \caption{\textbf{Mixamo Testset Characters.}}
  \label{fig:mixamo}
  \vspace{-0.4cm}
\end{figure}

\subsubsection{2D-to-3D (Synthetic) Comparison.} In order to quantitatively compare the motion retargeting performance with the previous methods, we compute the 2D-to-3D motion retargeting error on a synthetic animation dataset called Mixamo~\cite{mixamo} because it has ground-truths available. The training set comprises $32$ characters, and each character has $800$ actions. During evaluation, we test the models on a held-out partition with $4$ new characters (Fig.~\ref{fig:mixamo}) and $64$ new actions. Since 3DMPPE~\cite{Moon_2019_ICCV_3DMPPE} is not applicable to the 2D pose input, we further add two recent SOTA methods LCN~\cite{9174911} and Pose2Mesh \cite{Choi_2020_ECCV_Pose2Mesh} for the conventional pipeline. We further compare a strong benchmark where ground-truth 3D motion is retargeted with IK. It resembles the practice of using high-precision MoCap systems and serves as a lower bound of the error. The results are shown in Table~\ref{tab:mse}. Our approach outperforms previous methods by a considerable margin on all the test character pairs. The performance of \textit{3DGT+IK} (the upper-bound baseline with 3D pose grouth-truths) proves that the IK step itself is almost error-free. Hence, errors are mainly produced at the 3D stage and amplified in the IK stage as for the conventional two-stage pipeline. It is also notable that Solo-Dancer is only $30\%$ of Mixamo training set in terms of video length. The model trained on Solo-Dancer performs better, which demonstrates the strength of training on in-the-wild data with higher motion diversity.

\subsection{Canonicalization and Disentanglement}

By canonicalization training, the model learns to disentangle the three latent factors without external supervision. We first visualize the canonicalization results, which shows that the model learns to transform the skeleton sequences as intended. In addition, we show that the learned canonical form could serve as an effective representation of human motion.

\begin{figure}[t]
  \centering
  \includegraphics[width=\linewidth]{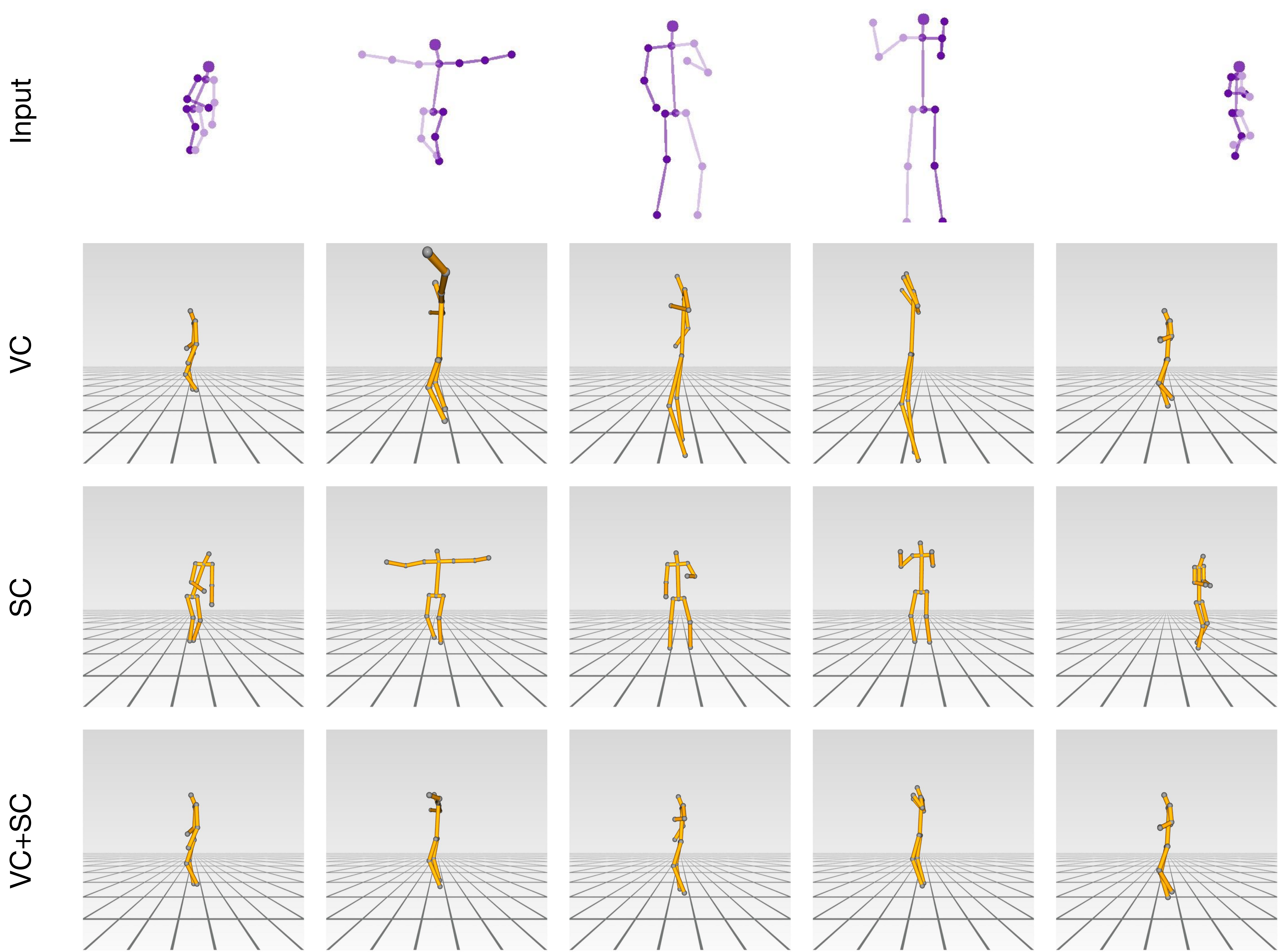}
  \caption{\textbf{Canonicalization Results.} The first row is the input 2D skeleton sequences, all from the Mixamo test set. The following rows show the 3D results after applying view canonicalization, structure canonicalization, and both.}
  \label{fig:scvc}
\end{figure}

\subsubsection{Canonicalization Effects.}

As shown in Fig.~\ref{fig:scvc}, input skeleton sequences with different orientations are aligned to the same view angle after view canonicalization (Row 2), while the motion and structure information are not changed. The view canonicalization requires the network to infer the underlying 3D structure of the input motion sequence and automatically cast it to a specific view. Despite the large input body structure discrepancies, structure canonicalization (Row 3) generates a ``standard character'' performing the same motion at the exact view angle. The structure canonicalization requires the network to capture the average structure and learn a character-agnostic motion representation. As shown in the last row of Fig.~\ref{fig:scvc}, the 3D skeleton sequence after both structure and view canonicalization contains unmixed motion information, independent of body structure and view angle. Therefore, it can be used as a direct, disentangled, and interpretable motion representation. We conduct further experiments to show the practicability of using the dual-canonicalized skeleton sequence as a distilled motion representation.

\begin{figure}[t]
  \centering
  \includegraphics[width=0.75\linewidth]{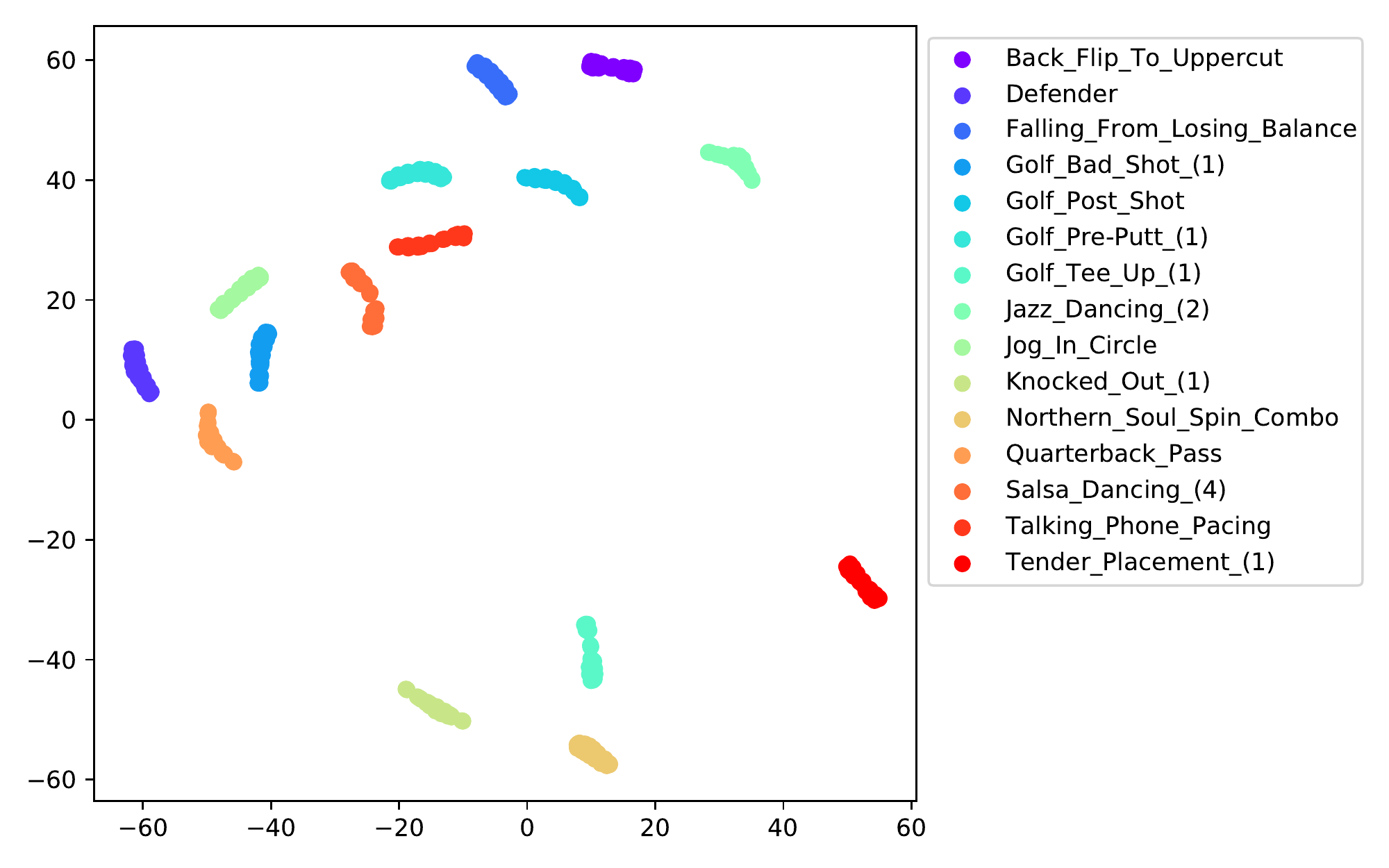}
  \caption{\textbf{Motion Clustering Visualization.} Motion of $4$ characters from $7$ view-angles are clustered after view canonicalization and structure canonicalization, and the top 10 clusters are plotted. All the clips are sampled from the Mixamo testset.}
  \label{fig:cluster}
\end{figure}

\begin{table}[t]
\caption{\textbf{Motion Clustering Evaluation.} For all the metrics, larger number means better clustering result.}
\begin{center}
\resizebox{\linewidth}{!}{
\small
\begin{tabular}{c|ccccc}
\Xhline{1.2pt}
 & ARI & AMI & Homogeneity & Completeness & V-Measure \\
\Xhline{1.2pt}
TransMoMo & 0.241 & 0.620 & 0.657 & 0.756 & 0.703 \\
Canonical & \textbf{0.347} & \textbf{0.707} & \textbf{0.750} & \textbf{0.808} & \textbf{0.778} \\
\Xhline{1.2pt}
\end{tabular}
}
\end{center}
\label{tab:clustering}
\end{table}

\begin{figure}[t]
  \centering
  \includegraphics[width=0.9\linewidth]{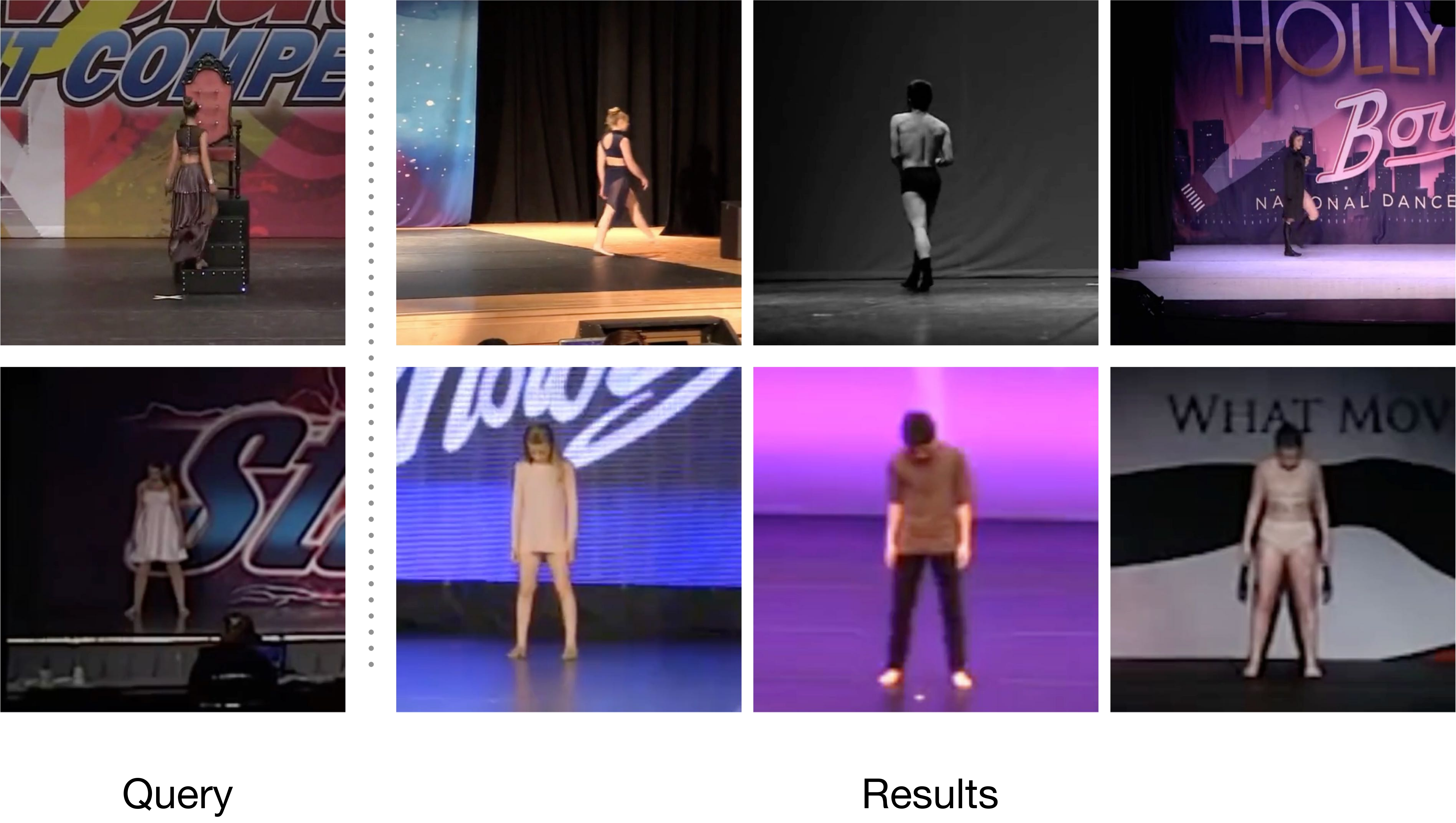}
  \caption{\textbf{Motion Retrieval Results.} The leftmost column shows a frame sampled from a query video depicting a short motion, and the three other columns show the top three retrieved videos.}
  \label{fig:retrieval}
\end{figure}

\subsubsection{Motion Clustering.} We apply a simple K-Means algorithm to cluster the Mixamo test set by motions. For K-Means, we set $K=64$ and iterate $300$ times. We visualize the motion clustering result in Fig. \ref{fig:cluster}. The test skeleton sequences are canonicalized and mapped to 2D using t-SNE. The network learns to disentangle and separate \textit{pure motion} through the proposed canonicalization operations despite the variance of character structure and camera view. The clustering results are further evaluated by ARI (Adjusted Rand index)~\cite{ARI}, AMI (Adjusted Mutual Information)~\cite{10.1145/280814.280820}, Homogeneity, Completeness, and V-Measure~\cite{rosenberg-hirschberg-2007-v}. We compare two motion representations: 1) Latent motion embedding learned by TransMoMo~\cite{yang2020transmomo}, and 2) Skeleton sequence after our canonicalization operations. Table~\ref{tab:clustering} shows that our canonical motion representation consistently outperforms the previous work. It suggests that the proposed canonicalization operations help to learn better motion representation.

\subsubsection{Motion Retrieval.} The dual-canonicalized motion representation also enables the motion retrieval task, \ie retrieve videos that exhibit similar motions with the query video. We first extract pose from the Solo-Dancer dataset, then slice the pose sequences into clips of $64$ frames, producing a motion library of $33678$ clips. Then, for a query motion, we retrieve its nearest neighbors in the dual-canonicalized motion representation space. The results in Fig.~\ref{fig:retrieval} show that our method is able to retrieve videos with similar motion semantics even if they appear drastically different due to body structure and view angle variations. It is because canonicalization operations make the proposed representation independent of both structure and view angle.

\subsubsection{Consistency Evaluation.}

\begin{figure}[h]
  \centering
  \includegraphics[width=\linewidth]{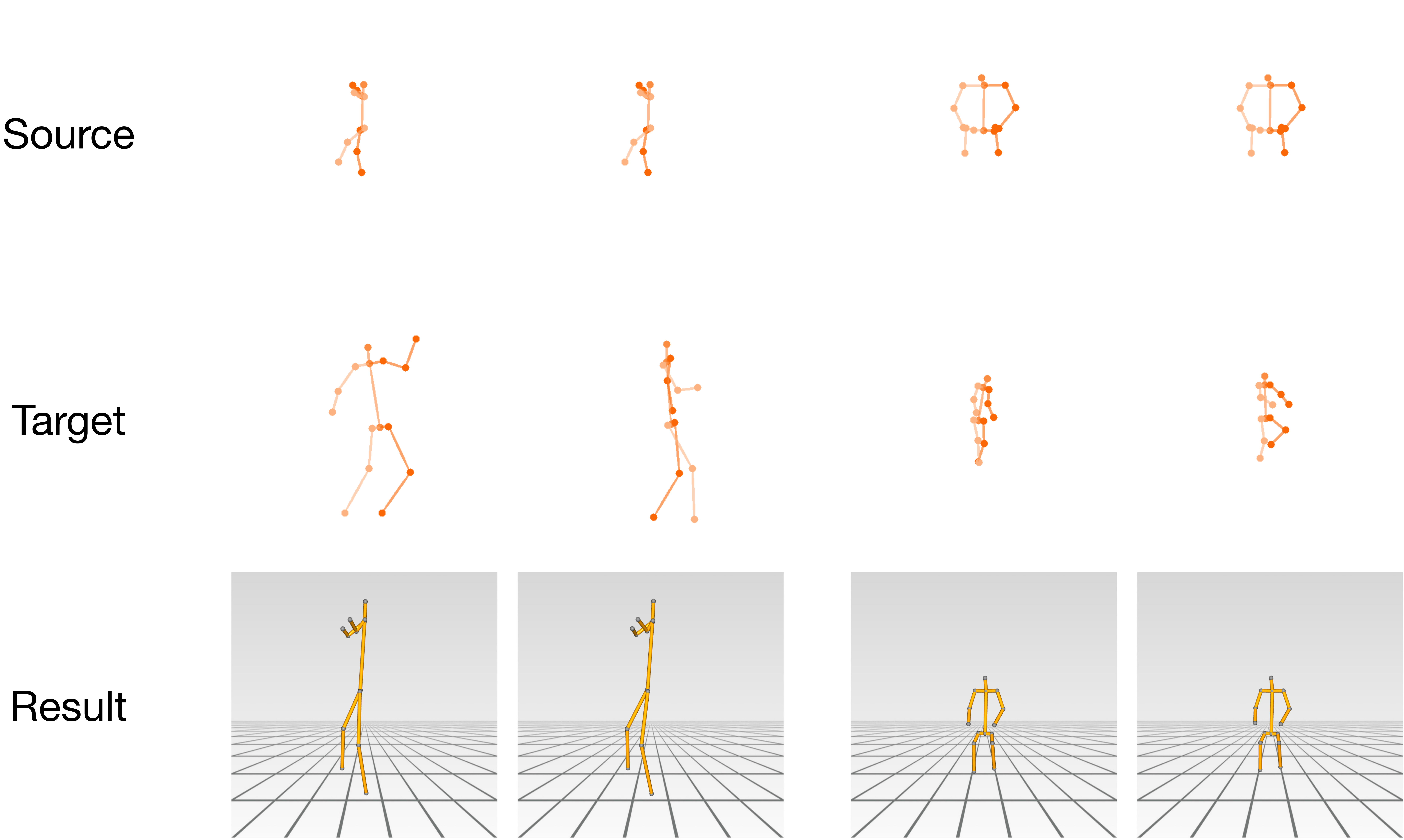}
  \caption{\textbf{Consistency Results.} For each group (left, right), the target character is the same but with different actions and viewpoints. Given the same source video, the retargeting results are consistent.
  }
  \label{fig:consistency}
\end{figure}

Ideally, the motion retargeting results should be robust to the variation of target character's motion and view-angle. To examine our method's capability of generating consistent motion-retargeted sequences, we test the same source sequence and target structure with varying target motion and view-angles, as shown in Fig. \ref{fig:consistency}.

\begin{table}[t]
\caption{\textbf{Ablation Study.} \textit{w/o} adv refers to the model with adversarial loss ablated, \textit{w/o} VC refers to the model with view canonicalization loss ablated, \textit{w/o} SC refers to the model with structure canonicalization loss ablated. T(3D) refers to TransMoMo re-implemented with 3D loss terms. T(3D+DV) refers to TransMoMo trained with 3D loss terms and dynamic view. The numbers are measured on the Mixamo dataset.} 
\begin{center}
\resizebox{\linewidth}{!}{
\small
\begin{tabular}{c|cccccc}
\Xhline{1.2pt}
 & \textit{w/o} adv & \textit{w/o} VC & \textit{w/o} SC & T(3D) & T(3D+DV) & Ours (Full) \\ \hline
\Xhline{1.2pt}
$\text{MSE} \times10^{-2}$ & 1.579 & 1.240 & 1.599 & 1.652 & 1.688 & \textbf{0.891} \\
$\text{MPJPE} \times10^{-1}$ &1.785 & 1.602 & 1.693 & 1.776 & 1.743 & \textbf{1.261} \\ 
\Xhline{1.2pt}
\end{tabular}
}
\end{center}
\label{tab:ablation}
\end{table}

\subsection{Ablation Study}

We first ablate several loss terms to show the importance of the designed modules. Then, we provide a stepwise comparison with TransMoMo~\cite{yang2020transmomo}. In Table~\ref{tab:ablation}, the removal of any part of our full model results in non-trivial performance degradation. More specifically, we observe that the network trained without adversarial loss tends to generate poses with a wrong depth, and the network trained without canonicalization loss fails to keep the desired motion. Although directly applying dynamic view and 3D loss terms to TransMoMo~\cite{yang2020transmomo} slightly helps, the models still exhibit a considerable performance gap compared to the proposed canonicalization-based approach. The models also fail to learn time-varying 3D view angle as expected. We argue that the proposed canonicalization operations not only help to learn disentangled motion better, but also ease the learning of dynamic view representations, which are more realistic yet more complex.

\section{Conclusion}

In this work, we present Canonicalization Networks for tackling in-the-wild 2D-to-3D motion retargeting. The proposed method robustly infers and transfers motion from 2D monocular videos to drive 3D humanoids, enabling motion retargeting in everyday use cases. We derive self-supervision from the canonicalization operations in factorized semantic spaces, from which the presented framework can be trained easily with web-crawled videos only given no annotations. To sum up, our 3D motion retargeting approach enables both training and inference in-the-wild. The experiment results show improvement of the proposed method over the existing methods, especially in terms of robustness and generalization.

\paragraph{Acknowledgement.} This study is partially supported under the RIE2020 Industry Alignment Fund Industry Collaboration Projects (IAF-ICP) Funding Initiative, as well as cash and in-kind contribution from the industry partner(s). We thank Xiaoxuan Ma, Shikai Li, Jiajun Su and Peizhuo Li for insightful discussion and kind support.
\clearpage

\begin{small}
\bibliography{egbib}
\end{small}

\FloatBarrier
\begin{table*}[t]
\begin{center}
\begin{tabular}{c|c|c|c|c|c|c|c|c|c|c|c|c|c|c|c|c}
~ & \multicolumn{5}{c|}{Layer 1} & \multicolumn{5}{c|}{Layer 2} & \multicolumn{5}{c|}{Layer 3} & Output \\
\hline
~ & c & k & p & s & pool & c & k & p & s & pool & c & k & p & s & pool & pool \\
\hline \hline
Motion & 64 & 8 & 3 & 2 & none & 128 & 8 & 3 & 2 & none & 128 & 8 & 3 & 2 & none & none\\
\hline
Structure & 64 & 7 & 2 & 1 & max & 128 & 7 & 2 & 1 & max & 256 & 7 & 2 & 1 & max & max\\
\hline
View & 64 & 7 & 3 & 1 & none & 32 & 7 & 3 & 1 & none & 6 & 7 & 3 & 1 & none & none\\
\hline
\end{tabular}
\end{center}
\caption{\textbf{Encoder Structure.} We use three layers of 1-D temporal convolution for our encoders. The parameter of the convolution on abbreviated: $c$ denotes number of output channels, $k$ denotes kernel size, $p$ denotes padding, $s$ denotes stride and $pool$ denotes the type of pooling. The pooling is also performed on the temporal dimension. }
\label{table:enc}
\end{table*}
\FloatBarrier

\section{Appendix}

\subsection{Network Architecture}
\paragraph{Encoders.} The sizes of the latent representations are: motion $C_m=128$, structure $C_s=256$. We use an explicit 6D representation \cite{zhou2019continuity} for view-angle, therefore $C_v=6$. The temporal dimension is down-sampled by the motion encoders to an eighth of the original length, therefore $M=\frac{8}{T}$. The detailed network structure is shown in Table \ref{table:enc}.

\paragraph{Decoder.} The decoder is also a three-layer temporal convolutional network. The input channel is $C_m + C_s = 384$ and the number of output channels of each layer are $256$, $128$ and $45$. The kernel size is $7$.

\subsection{Evaluation Metrics}

For an inferred sequence $\hat{\mathbf{X}}$ and a groundtruth sequence $\mathbf{X}$,
\[\text{MSE} = \frac{1}{3NT} \sum_{i, t} \frac{ \left \| \mathbf{X}_{i,t} - \hat{\mathbf{X}}_{i, t} \right \|^2 }{h^2(\mathbf{X})} \]
\[\text{MPJPE} = \frac{1}{NT} \sum_{i, t} \frac{\left \| \mathbf{X}_{i,t} - \hat{\mathbf{X}}_{i, t} \right \|}{h(\mathbf{X})} \]
where $i$ is the subscript of body joints and $t$ is the subscript of time; $h(\cdot)$ calculates the height of a character by summing the lengths of the head, body and legs. This normalization is introduced to offset the effect that errors tend to be larger on larger characters. The errors are computed after hip-alignment so that the measurement is focused on the movements of different body parts local to the hip joint.

\end{document}